# A Genetic Algorithm for solving Quadratic Assignment Problem (QAP)

H. Azarbonyad[a], R. Babazadeh[b]

[a]Department of Electrical and computers engineering, University of Tehran, Tehran, Iran, h.azarbonyad@ece.ut.ac.ir
[b]Department of industrial engineering, University of Tehran, Tehran, Iran, r.babazadeh@ut.ac.ir

*Abstract*— **The Quadratic Assignment Problem (QAP) is one of the models used for the multi-row layout problem with facilities of equal area. There are a set of *n* facilities and a set of *n* locations. For each pair of locations, a distance is specified and for each pair of facilities a weight or flow is specified (e.g., the amount of supplies transported between the two facilities). The problem is to assign all facilities to different locations with the aim of minimizing the sum of the distances multiplied by the corresponding flows. The QAP is among the most difficult NP-hard combinatorial optimization problems. Because of this, this paper presents an efficient Genetic algorithm (GA) to solve this problem in reasonable time. For validation the proposed GA some examples are selected from QAP library. The obtained results in reasonable time show the efficiency of proposed GA.**

*Key words*- **Genetic Algorithm, QAP, Multi-row layout problems**

## 1. INTRODUCTION

Some multi-row layout problems are the control layout problem, the machine layout problem in an automated manufacturing system and office layout problem also the Travelling Salesman Problem (TSP) may be seen as a special case of QAP if one assumes that the flows connect all facilities only along a single ring, all flows have the same non-zero (constant) value. Many other problems of standard combinatorial optimization problems may be written in this form (see [1] and [2]).

The Quadratic Assignment Problem (QAP) is one of the classical combinatorial optimization problems and is known for its diverse applications and is widely regarded as one of the most difficult problem in classical combinatorial optimization problems. The QAP is NP-hard optimization problem [7], so, to practically solve the QAP one has to apply heuristic algorithms which find very high quality solutions in short computation time. And also there is no known algorithm for solving this problem in polynomial time, and even small instances may require long computation time [2].

The location of facilities with material flow between them was first modeled as a QAP by Koopmans and Beckmann [3]. In a facility layout problem in which there are $n$ facilities to be assigned to $n$ given location. The QAP formulation requires an equal number of facilities and locations. If there are fewer than n, say $m<n$, facilities to be assigned to $n$ locations, then to use the QAP formulation, $n$-$m$ dummy facilities should be created and a zero flow between each of these and all others must be assigned ( including the other dummy facilities). If there are fewer locations than facilities, then the problem is infeasible [4]. Considering previous works in QAP ([2],[8] and [9]) this paper is developed to solve this problem with GA in short computation time.

In the next section the QAP is described and formulated. After introducing GA in Section 3, the encoding scheme, solution representation and GA operators are described in this section. The computational results are reported in Section 4. Finally, Section 5 concludes this paper and offers some directions for further research.

## 2. Mathematical model

The following notation is used in formulation of QAP [4]:

*parameters*

    $n$    total number of facilities and locations
    $f_{ik}$    flow of material from facility *I* to facility *k*
    $d_{jl}$    distance from location *j* to location *l*

*Variable*

$$x_{ij} = \begin{cases} 1 & \text{If facility } I \text{ is assigned to location } j \\ 0 & \text{Otherwise} \end{cases}$$

$$\text{Min} \quad \sum_{i}^{n} \sum_{j}^{n} \sum_{k}^{n} \sum_{l}^{n} f_{ik}\, d_{jl}\, x_{ij}\, x_{kl} \tag{1}$$

$$\sum_{j}^{n} x_{ij} = 1 \quad \forall i \tag{2}$$

$$\sum_{i}^{n} x_{ij} = 1 \quad \forall j \tag{3}$$

$$x_{ij} \text{ is binary} \tag{4}$$

The objective function (1) minimizes the total distances and flows between facilities. Constraints (2) and (4) ensure that each facility *I* is assigned to exactly one location. As well as Constraints (3) and (4) assure that each location *j* has exactly one facility which assigned to it. The term quadratic stems from the formulation of the QAP as an integer optimization problem with a quadratic objective function.

## 3. Genetic Algorithm

Genetic Algorithms (Gas) are routinely used to generate useful solutions to optimization and search problems. Genetic algorithms belong to the larger class of evolutionary algorithms (EA), which generate solutions to optimization problems using techniques inspired by natural evolution. In a genetic algorithm, a population of strings (called chromosomes or the genotype of the genome), which encode candidate solutions (called individuals, creatures) to an optimization problem, evolves toward better solutions. Traditionally, solutions are represented in binary as strings of 0s and 1s, but other encodings are also possible. The evolution usually starts from a population of randomly generated individuals and happens in generations. In each generation, the fitness of every individual in the population is evaluated, multiple individuals are stochastically selected from the current population (based on their fitness), and modified (recombined and possibly randomly mutated) to form a new population. The new population is then used in the next iteration of the algorithm. Commonly, the algorithm terminates when either a maximum number of generations has been produced, or a satisfactory fitness level has been reached for the population. If the algorithm has terminated due to a maximum number of generations, a satisfactory solution may or may not have been reached [5].

### 3.1. Encoding scheme

In this section we describe the encoding scheme for different component of proposed GA when n=5. For creating initial population we consider a chromosome with n gene, as any gene is depictive of assignment each facility to exactly one location. For example the chromosome in Figure 1 show that facilities 2,4,3,1 and 5 is assigned to locations 1,2,3,4 and 5 respectively that it is a feasible solution.

| 2 | 4 | 3 | 1 | 5 |
|---|---|---|---|---|
| 1 | 2 | 3 | 4 | 5 |

Fig.1. chromosome representation

For making above chromosome (feasible solution) to create initial population, firstly we assume that all content of chromosome are placed equal to 1, then random integer number from 1 to 5 is produced, say that 3, and then it is compared with the content of corresponding square, so if the corresponding content is equal to 1, facility 3 is assigned to location 1 and then this content is set equal to zero (see figure 2). Again a random integer number from 1 to 5 is produced, then If it is equal to 3 because of the corresponding content is zero (it means that facility 3 is assigned) so, another number should be produced. This way continues until all facilities are assigned to locations and also the determined number of initial population of chromosomes is satisfied.

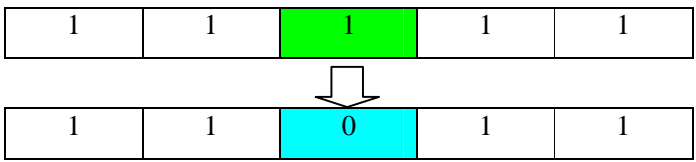

Fig.2. making chromosome (feasible solution)

### 3.1.1. Crossover

We use two point crossovers for implementation of GA. For making child 1 and child 2, two points are randomly selected from 1 to 5 and between these points is hold fixed. Then, child 1 is created according to orders of parent 2 and fixed components of parent 1, as well as child 2 is produced according to order of parent 1 and fixed components of parent 2 (see Figure 3). After tuning the crossover probability is determined equal to 0.8. This means that 80% of selected parents participate in crossover operation.

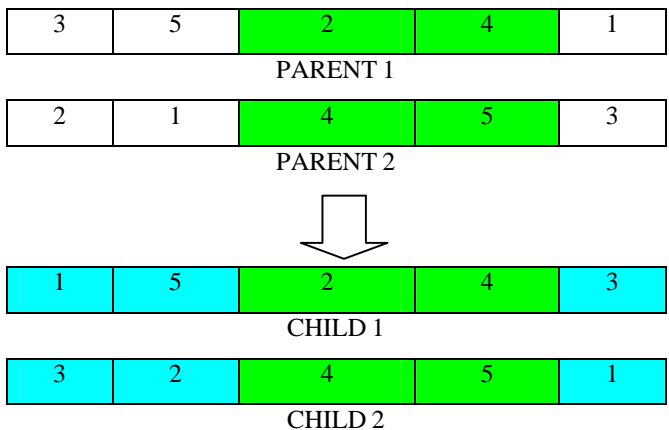

FIG.3. CROSSOVER REPRESENTATION

### 3.1.2. Mutation

Mutation operator changes the value of each gene in a chromosome with probability 0.2. For mutation two genes are randomly selected from selected chromosome and then their locations are substituted (see Figure 4).

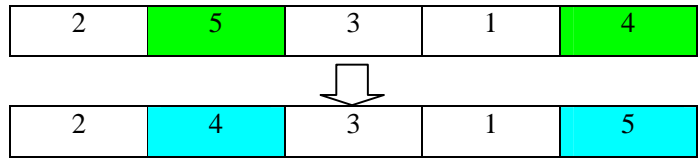

Fig.4. Mutation representation

### 3.1.3. Selection

In selection operator the number of chromosomes required to complete next generation from crossover, mutation and previous generation are determined special mechanism that we use roulette wheel mechanism. In roulette wheel selection, candidate solutions are given a probability of being selected that is directly proportionate to their fitness. The selection probability of a candidate solution $I$ is $\dfrac{F_i}{\sum_{j}^{P} F_j}$, where $F_i$ is the fitness of chromosome $I$ (objective function in QAP formulation) and $P$ is the population size.

## 4. Computational Results

For validation the proposed GA some examples are selected from QAP library. The proposed GA was programmed in Java net beans 6.9.1 on a Pentium dual-core 2.66 GHZ computer with 4 GB RAM. The objective function values of the best known solutions are given by Burkard et al. [6].

As it is illustrated in Table 1 the proposed GA results in the short computational time with acceptable GAP.

**Table 1:** Computational results of proposed GA

| Test Problems | Global/local optimum | Gap% | Run time (second) |
|---|---|---|---|
| Nug 12 | global | 0 | 1 |
| Nug 17 | local | .0034 | 3 |
| Nug 20 | local | 0 | 3 |
| Nug 24 | local | .0034 | 4 |
| Nug 28 | local | .012 | 5 |
| chr12a.dat | local | 0 | 1 |
| chr12b.dat | local | 0 | 2 |
| chr15a.dat | local | 1 | 4 |

## 5. Conclusion

In this paper at first QAP as a special type of multi-row layout problem with facilities of equal area is introduced and its various applications in different fields are described. Then, due to NP-hard nature of QAP an efficient GA algorithm is developed to solve it in reasonable time. Finally, some computational results from data set of QAP library are provided to show the efficiency and capability of proposed GA. Comparing outcome results with those works available in literature justify the proficiency of proposed GA in achieving acceptable solutions in reasonable time. For future research direction, developing other heuristic and/or meta-heuristic such as simulated annealing (SA) algorithm and PSO algorithm and then comparison results of them with results of this paper can be attractive work.